# Lung Cancer Detection Using Deep Learning

Imama Ajmi[1], Abhishek Das[2*]

[1]Department of Computer Sc. & Engg., B. P. Poddar Institute of Management and Technology, Kolkata, India

[2]Department of Computer Sc. & Engineering, Aliah University, Kolkata, India

*Corresponding Author Email: adas@aliah.ac.in

**Competing Interest Statement**

The authors declare that they have no known competing financial interests or personal relationships that could have appeared to influence the work reported in this paper.

**Funding Statement**

This research did not receive any specific grant from funding agencies in the public, commercial, or not-for-profit sectors.

# Lung Cancer Detection Using Deep Learning

## Abstract

Lung cancer, the second leading cause of cancer-related deaths, is primarily linked to long-term tobacco smoking (85% of cases). Surprisingly, 10-15% of cases occur in non-smokers. In 2020, approximately 2 million people were affected globally, resulting in 1.5 million deaths. The survival rate, at around 20%, lags behind other cancers, partly due to late-stage symptom manifestation. Necessitates early and accurate detection for effective treatment. Performance metrics such as accuracy, precision, recall (sensitivity), and F1-score are computed to provide a comprehensive evaluation of each model's capabilities. By comparing these metrics, this study offers insights into the strengths and limitations of each approach, contributing to the advancement of lung cancer detection techniques. In this paper, we are going to discuss the methodologies of lung cancer detection using different deep learning algorithms - InceptionV3, MobileNetV2, VGG16, ResNet152 - are explored for their efficacy in classifying lung cancer cases. Our Proposed Model algorithm based is a 16 layers architecture based on CNN model. Our Proposed model exhibits several key highlights that contribute to its novelty. By integrating multiple layer types such as convolutional, pooling, flatten, dropout, fully connected and dense layers, the model leverages the strengths of each layer to enhance its predictive capabilities. Novelty of our proposed model is that its accuracy is increasing consistently with the increasing no of epochs. We have tested the model performance up to epoch no 30. Our proposed model also overcome the overfitting problem.

## 1. Introduction

In 2020, approximately 2 million people were affected globally, resulting in 1.5 million deaths. The survival rate, at around 20%, lags behind other cancers, partly due to late-stage symptom manifestation [26][28].

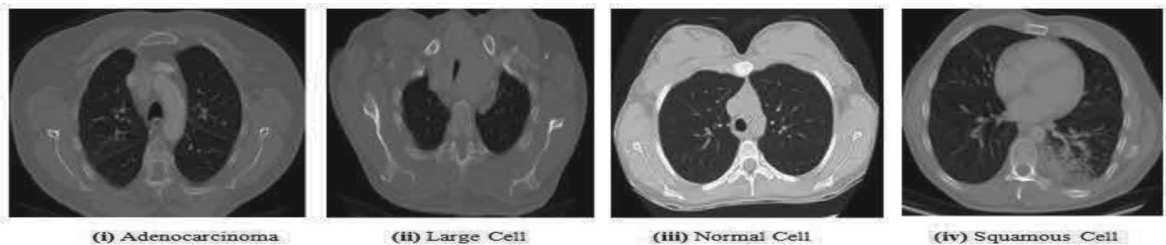

**Fig. 1:** Images of 4 Classes of the Dataset

In this work, I have taken an chest CT scan image dataset from kaggle as input. Then I have used five different CNN models by increasing the hidden layers respectively. Also build a Proposed Model to classify. After that I have checked the difference between the outputs of the different CNN models mainly the prediction accuracy. I have also calculated precision, recall, F1-score of each model.

If clinical experts and technicians have lack of knowledge and skills, the experimental result may be wrong. And also the instrument which has been used in the clinic for experiment that instrument maybe old version /model and most of the instrument are maybe damage for lack of regular cleaning and maintenance. And the most serious problem is the clinical experts are not getting their required lab for work. Above all this problem patient are not getting their accurate result .Most of the time patients get false positive and false negative results for that they are suffering a lot of problem, like if any patient has not lung cancer but result comes false positive for that doctor gives the patient anti-cancer medicine and radio therapy after taking those medicines and treatments patient suffer different types of side effects. False negative results are also dangerous for patient because patient has cancer but the result comes negative the doctor would not treatment the patient. For that cancer cells will grow in the patient body day by day, which is very fatal for the patient. In this project we are trying to reduce this type of problem and gives accurate result for the patient. [25].

Overall, the proposed model exhibits novelty in its effective utilization of convolutional layers, integration of various layer types, consideration of feature extraction and dimensionality reduction capabilities, and interpretability. These qualities make it a promising approach for lung cancer detection, offering the potential to deliver accurate and interpretable predictions. With further research and validation, the proposed model holds promise for advancing the field of lung cancer detection using deep learning and ultimately contributing to improved healthcare outcomes.

## 2. Review of literature

The paper [1] introduces a CNN model for detecting lung cancer from thoracic CT scans. The dataset consisted of thousands of CT images. Data augmentation and transfer learning were utilized to improve model performance. The CNN architecture included multiple convolutional layers for feature extraction. An accuracy of 95% was achieved. The use of transfer learning helped mitigate overfitting. Sensitivity and specificity metrics were also favorable. The model demonstrated significant potential for early lung cancer detection. Key techniques included advanced feature extraction and classification. The study addressed common challenges in medical imaging. The authors emphasized the importance of high quality training data. The model's robustness was tested on different datasets. Performance metrics showed promising results. The study highlights the importance of early intervention. Clinical implications were discussed. Future research directions were suggested.

The paper [2] systematic reviews and explores various deep learning architectures for lung cancer diagnosis. Models such as VGG-16, ResNet, and Inception were analyzed. The review highlights advancements in pulmonary nodule detection. Multimodal data integration was a key focus. High sensitivity and specificity rates were reported. The strengths and limitations of different models were discussed. Comprehensive datasets were deemed crucial for model training. The study emphasizes the role of data preprocessing. Results from different studies were compared. The review provides insights into clinical applications. Challenges in implementing deep learning in healthcare were addressed. Future research directions were identified. The review concludes that deep learning significantly enhances diagnostic accuracy. It also emphasizes the need for interdisciplinary collaboration. Clinical applicability and potential improvements were discussed.

The paper [3] developed a deep learning model for early-stage lung cancer detection using CT images. The model achieved a prediction accuracy of 97%. Early detection capabilities were a major focus. Advanced image preprocessing techniques were employed. Data augmentation helped improve model robustness. The model demonstrated high precision in identifying early cancer signs. Sensitivity and specificity metrics were favorable. The study underscores the importance of early intervention. Clinical implications were discussed. Overfitting issues were addressed through regularization techniques. The approach showed potential for real-world applications. Performance metrics were promising. Future improvements were suggested.

This paper [4] explores deep feature extraction combined with ensemble learning for cancer detection. Both lung and colon cancers were addressed. The hybrid model achieved high accuracy. Large datasets were used for training and validation. Advanced feature extraction techniques were employed. Ensemble learning improved model robustness. The study highlights the benefits of multimodal data integration. Results demonstrated significant diagnostic potential. Computational efficiency was also considered. Data preprocessing played a crucial role. Sensitivity and specificity metrics were favorable. The approach was validated on clinical data. The study underscores the importance of comprehensive datasets. Clinical applicability is emphasized. Future research directions were suggested.

The paper [5] introduces DFD-Net, focusing on denoising CT scan images before detection. The model significantly improved accuracy by reducing noise. Enhanced image clarity was achieved. Lung nodule detectability was improved. The importance of preprocessing was highlighted. Results showed high accuracy metrics. Validation was performed on multiple datasets. Computational requirements were optimized. The study addresses common challenges in medical imaging. Performance metrics were promising. The model's robustness was tested extensively. Sensitivity and specificity were favorable. The research underscores the role of image quality. Clinical implications were discussed. Future improvements were suggested.

The paper [6] used Advanced CNN architectures. The model first detects cancer and then stages it based on severity. High sensitivity and specificity were achieved. The approach aids in treatment planning. Robust feature extraction techniques were employed. Validation was performed on clinical data. The study emphasizes comprehensive cancer management. Results showed significant diagnostic accuracy. Data augmentation helped improve model performance. Overfitting was addressed through regularization. Performance metrics were favorable. The study focuses on early detection and accurate staging. Clinical implications were discussed. Future research directions were suggested.

This paper [7] did comparative study evaluates various DL networks to detect lung cancer. CNNs and RNNs were primarily considered. The paper focuses the superior performance of DL models over traditional methods. Hybrid CNN-RNN models showed the highest accuracy. Data augmentation techniques were used. Overfitting issues were addressed. Results demonstrated high diagnostic potential. Sensitivity and specificity were favorable. Validation was performed on large datasets. The study underscores the importance of model comparison. Computational efficiency was also considered. Performance metrics were promising. The research highlights the need for continuous improvement. Clinical applicability was discussed. Future research directions were suggested.

The paper [8] presents a modified Inception Recurrent RCNN for lung carcinoma diagnosis. Histopathological images were used. The model achieved high accuracy and efficiency. Data preprocessing techniques were employed. Feature extraction was enhanced. Validation was performed on a reduced dataset. Early cancer detection capabilities were a major focus. Performance metrics were promising. Sensitivity and specificity were favorable. The study highlights the importance of advanced image processing. Clinical implications were discussed. Overfitting issues were addressed through regularization techniques. Results showed significant diagnostic potential. The research underscores the importance of early intervention. Future research directions were suggested.

This paper [9] develops a DL model for lung cancer detection from chest X-rays. CNNs were used for feature extraction. High accuracy rates were achieved. Large-scale screening potential is emphasized. Advanced preprocessing techniques were employed. Validation was performed on large datasets. Results demonstrated significant diagnostic accuracy. Sensitivity and specificity were favorable. The study addresses overfitting issues. Regularization techniques were used. The model's robustness was tested extensively. Performance metrics were promising. The research underscores the importance of chest X-ray imaging. Clinical applicability is emphasized. Future improvements were suggested.

This paper [10] combines the Tuna Swarm with DL for cancer detection. Both lung and colon cancers were addressed. High accuracy and reduced complexity were achieved. Advanced preprocessing techniques were employed. Large datasets were used for training and validation. Results demonstrated significant diagnostic potential. Computational efficiency was optimized. Sensitivity and specificity metrics were favorable. The study focuses the multimodal data integration. Clinical implications were discussed. Overfitting issues were addressed. The model's robustness was tested extensively. Performance metrics were promising. The research underscores the role of advanced algorithms. Future research directions were suggested.

The paper [11] presents a multimodal fusion approach. CT images and clinical data were integrated. A deep neural network was used. Advanced preprocessing techniques were employed. Results showed significant diagnostic accuracy. The approach enhances robustness. Large datasets were used for validation. The study underscores the importance of comprehensive data integration. Sensitivity and specificity metrics were favorable. Performance metrics were promising. Overfitting issues were addressed through regularization. Clinical applicability was emphasized. The research highlights the potential for real-world applications. Future improvements were suggested.

The paper [12] reviews and analyzes advancements in DL methods for lung cancer. Various architectures, including CNNs and RNNs, were considered. Advancements in feature extraction techniques were highlighted. Multimodal data integration was a key focus. High sensitivity and specificity rates were reported. Strengths and limitations of different models were discussed. Comprehensive datasets were crucial for training robust models. Preprocessing techniques were vital. Results from multiple studies were compared. Clinical applications were explored. Implementation challenges were addressed. Future research directions were outlined. Clinical implications and potential improvements were discussed.

The paper [13] proposes a compound DL model for lung cancer. The model combines CNN and LSTM architectures. High accuracy and efficiency were achieved. Data augmentation techniques were employed. Large datasets were used for training. Advanced feature extraction methods were implemented. Sensitivity and specificity metrics were favorable. Validation was performed on clinical datasets. Robustness was a key focus.

Overfitting was addressed through regularization. The paper highlights the importance of compound models. Clinical applicability was discussed. Performance metrics included precision, recall, and F1-score. Future research directions were suggested. The study emphasizes importance of hybrid models.

## 3.  Methodology

As per our requirement, we are going to build a deep neural network model to solve the previous problems in the existing model and improve the accuracy and efficiency. Designing an accurate and reliable model for lung cancer detection is crucial in the healthcare domain. Deep learning has emerged as a powerful approach for handling complex medical data and extracting meaningful patterns. In this study, we propose a novel deep learning (CNN-Based) model specifically tailored for lung cancer detection. Our model is designed to leverage the strengths of deep learning techniques.

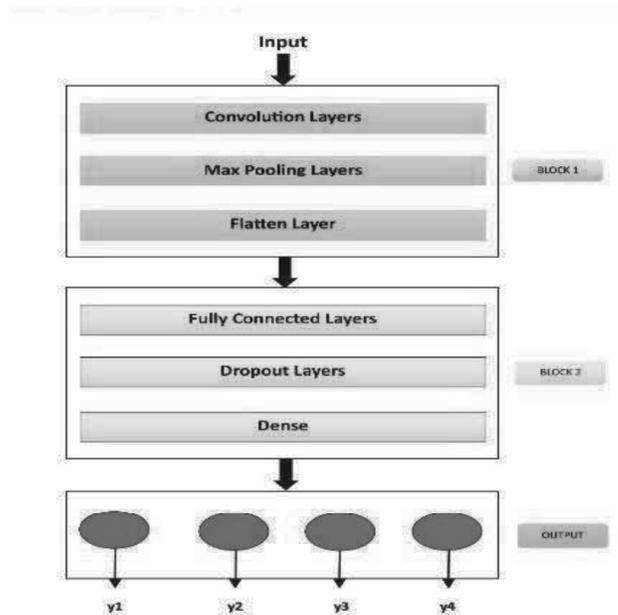

**Fig. 2:** Flowchart of Proposed Model

I choose this architecture for chest CT scan image analysis because it progressively reduces spatial dimensions while retaining important features, enhancing pattern detection and robustness. The 2D convolutional layers with ReLU activation capture spatial hierarchies, and max-pooling reduces dimensionality and computational load. Repeating this combination five times ensures deep feature extraction. The flatten layer converts the 2D feature maps to a 1D feature vector. The flatten layer transitions to fully connected layers for high-level reasoning. Adding dropout layers prevents overfitting. Repeat this twice for robust feature learning. And the final dense layer with Softmax activation function provides classification probabilities for accurate diagnosis.

In certain situations, the custom model shown in the flowchart (Fig. 2) might be superior to a conventional, pre-built CNN model for a number of reasons.

1. Task-certain Architecture: multi-output regression/classification with $y_1$ to $y_4$ is an example of an architecture that is tailored for a certain application. Built-in models (such as VGG, ResNet, etc.) are often ImageNet and are intended for generic picture classification. Such customized models can concentrate on a specific area, increasing precision and effectiveness.

2. Lighter and More Efficient: For small or medium datasets, this model is more computationally efficient since it eliminates undue depth or complexity. Millions of parameters make for pretrained models, which frequently overkill for tasks particular to a given area.

3. Custom convolution, pooling, and flattening layers tailored to the input size are used in Block 1 of the layer customization process, while fully linked + dropout layers are used for regularization in Block 2. Because of its adaptability, the model can be accurately tuned, something that fixed prebuilt layers make difficult.

4. Dropout Layers: Specifically designed to avoid overfitting, dropout is particularly useful in smaller datasets where the huge capacity of pretrained models may cause them to overfit.

5. Support for Multiple Outputs: Multi-label or multi-output learning is indicated by the output layer's multiple predictions ($y_1$ to $y_4$). Only single-label categorization is supported by the majority of built-in models, necessitating considerable adjustments.

6. Improved Interpretability: It is simpler to debug and interpret a custom model.Unlike opaque blocks in prebuilt models, you are aware of the functions of each layer.

## 4. Result

**Table 1:** Summary table to compare all the used models

| Model Name | Epoch No. | Train Loss | Train Acc | Val Loss | Val Acc | Test Loss | Test Acc | Precision | Recall | F1 Score | RMSE | Acc Score |
|---|---|---|---|---|---|---|---|---|---|---|---|---|
| **Inception V3** | 10 | 0.6493 | 0.8504 | 0.6929 | 0.8358 | 9.2560 | 0.3651 | 0.44 | 0.45 | 0.39 | 1.4265 | 0.3651 |
| | 20 | 0.8626 | 0.8267 | 0.8621 | 0.7985 | 11.9857 | 0.4571 | 0.51 | 0.53 | 0.47 | 1.7229 | 0.4571 |
| | 30 | 0.7596 | 0.8770 | 0.6689 | 0.8731 | 12.4272 | 0.4286 | 0.49 | 0.55 | 0.45 | 1.4939 | **0.4286** |
| | | | | | | | | | | | | |
| **MobileNet V2** | 10 | 0.4129 | 0.8815 | 0.4198 | 0.8582 | 0.9290 | 0.5841 | 0.65 | 0.66 | 0.62 | 1.3916 | 0.5841 |
| | 20 | 0.3159 | 0.8993 | 0.3235 | 0.8955 | 1.2249 | 0.5206 | 0.67 | 0.62 | 0.55 | 1.3464 | 0.5206 |
| | 30 | 0.1971 | 0.9600 | 0.2233 | 0.9403 | 0.9228 | 0.6127 | 0.65 | 0.69 | 0.64 | 1.2574 | **0.6127** |
| | | | | | | | | | | | | |
| **VGG16** | 10 | 0.4714 | 0.8133 | 0.4403 | 0.8209 | 1.3323 | 0.6254 | 0.63 | 0.66 | 0.63 | 1.2737 | 0.6254 |
| | 20 | 0.2159 | 0.9348 | 0.2157 | 0.9403 | 0.8649 | 0.6794 | 0.69 | 0.70 | 0.69 | 1.2742 | 0.6794 |
| | 30 | 0.1248 | 0.9704 | 0.1162 | 0.9701 | 0.7083 | 0.7522 | 0.76 | 0.75 | 0.75 | 1.1533 | **0.7522** |
| | | | | | | | | | | | | |
| **ResNet152** | 10 | 0.2287 | 0.9407 | 0.5643 | 0.7556 | 0.5643 | 0.7556 | 0.77 | 0.80 | 0.78 | 1.1561 | 0.7556 |
| | 20 | 1.0216 | 0.7733 | 1.1359 | 0.7910 | 1.6644 | 0.6508 | 0.78 | 0.67 | 0.62 | 1.4387 | 0.6508 |
| | 30 | 0.0718 | 0.9985 | 0.0710 | 1.000 | 0.5303 | 0.7778 | 0.78 | 0.81 | 0.79 | 1.0954 | **0.7778** |
| | | | | | | | | | | | | |
| **Proposed Model** | 10 | 0.5242 | 0.8044 | 0.4913 | 0.8060 | 0.6794 | 0.6349 | 0.67 | 0.70 | 0.67 | 1.4606 | 0.6349 |
| | 20 | 0.0459 | 0.9881 | 0.0331 | 0.9925 | 0.9572 | 0.7079 | 0.73 | 0.76 | 0.73 | 1.3117 | 0.7079 |
| | 30 | 0.0088 | 0.9985 | 0.0032 | 1.000 | 0.6405 | 0.8444 | 0.85 | 0.87 | 0.86 | 0.9562 | **0.8444** |

The results for CNN model InceptionV3 with convolutional layer and max pool layer gives Training accuracy 87.70% and Test accuracy is 42.86% on the epoch no 30. Dataset is same for all five CNN model in this project . The result of deep learning CNN model MobileNetV2 with convolutional layer and MaxPooling layer gives Training accuracy 96% and Test accuracy is 61% with epoch number of 30. The result for CNN model VGG16 with convolutional layer and maxpool layer gives Training accuracy 97.04% and Test accuracy is 75.24% with the epoch no of 30. VGG16 achieving the test accuracy 62.54% with epoch number of 10, which is even better than previous two models (InceptionV3, MobileNetV2) achieving at epoch number 30. The result for the model of CNN is ResNet152 with convolutional layer and MaxPooling layer gives Training accuracy 99.85% and Test accuracy 77.78% with the epoch no of 30. ResNet152 model is achieving test accuracy 75.56%, which is also better result than previous three models including VGG16 at epoch number 10. Among these four inbuilt CNN model ResNet152 is providing best result. But their accuracy is not improving always with the number of epochs increment. The result for our Proposed Model with various types of layer gives Training accuracy 99.85% and Test accuracy is 84.44% with epoch no of 30, which is providing best result among all these 5 models. The result among these five CNN models - InceptionV3, MobileNetV2, VGG16, ResNet152 and our Proposed Model adding more epoch Training and Test accuracy are increased in the model VGG16 and Proposed Model but InceptionV3, MobileNetV2, and ResNet152 in these three models adding more epochs sometimes also decreased Training accuracy and Testing accuracy in the dataset of Chest CT Scan images. So here our Proposed Model is providing best result and also constantly improved accuracy with no of epoch increment. Also computation time is lesser than ResNet152, which is an inbuilt CNN model provides second best result with Chest CT Scan Kaggle dataset.

## 5. Conclusion

We all know that lung cancer is a life threatening disease. Our proposed model for lung cancer detection using deep learning possesses several notable qualities that contribute to its novelty. In our proposed model we have used 5 2D convolution layers with ReLU activation function, and after every Convolution layer there is a MaxPooling layer. So, total no of MaxPooling layer is also 5. Flatten layer is also applied after those 10 layers. Then we have used 2 Fully Connected Layers again with ReLU function, and each layer is followed by a dropout layer. Lastly Dense Layer with Softmax activation function is added. By incorporating these layer types, the model can effectively learn intricate patterns and enhance its predictive accuracy. Additionally, the model integrates various layers such as convolutional, pooling, flattening, and fully connected layers, harnessing the strengths of each layer type to optimize its predictive capabilities. Our Proposed Model is also providing the best result among all the models we have used in this project. Furthermore, the model incorporates feature extraction and dimensionality reduction techniques through convolutional and max pooling layers. These techniques help in identifying relevant features and reducing the overall data dimensionality, which can enhance the model's prediction accuracy. Novelty of the proposed model is consistency in the accuracy both for training and testing dataset. And we have examined it up to epoch no 30. For domain-specific tasks, this custom CNN model is probably superior to built-in CNN models. Deployment that is lightweight, interpretable, multi-output classification and regression, scenarios for small/medium datasets

Overall, the novelty of the proposed model lies in its effective utilization of convolutional and integration of various layer types, consideration of feature extraction and dimensionality reduction capabilities, and interpretability. These qualities make the proposed model a promising approach in the field of lung cancer detection using deep learning, with the potential to provide accurate and interpretable predictions. In future, we will improve the accuracy level of our model by tuning the sample data. It's crucial to address ethical and legal considerations in the development and deployment of these models, ensuring privacy, fairness, and transparency. Conducting clinical validation studies is essential to validate model efficacy, safety, and real-world applicability, bridging the gap between research and practical healthcare implementations.